\documentclass{bmvc2k}

\title{Motion Segmentation Using Locally \\ Affine Atom Voting}

\addauthor{Erez Posner}{erezposner@gmail.com}{1}
\addauthor{Rami Hagege}{ramih1000@gmail.com}{2}

\addinstitution{
Ben-Gurion University\\
 of the Negev
}
\addinstitution{
Ben-Gurion University\\
 of the Negev
}

\runninghead{}{}

\begin{document}

\maketitle

\begin{abstract}
We present a novel method for motion segmentation called LAAV (Locally Affine Atom Voting). Our model's main novelty is using sets of features to segment motion for all features in the scene. LAAV acts as a pre-processing pipeline stage for features in the image, followed by a fine-tuned version of the state-of-the-art Random Voting (RV) method. Unlike standard approaches, LAAV segments motion using feature-set affinities instead of pair-wise affinities between all features; therefore, it significantly simplifies complex scenarios and reduces the computational cost without a loss of accuracy. We describe how the challenges encountered by using previously suggested approaches are addressed using our model. We then compare our algorithm with several state-of-the-art methods. Experiments shows that our approach achieves the most accurate motion segmentation results and, in the presence of measurement noise, achieves comparable results to the other algorithms.

\end{abstract}

\section{Introduction}
\label{sec:intro}
A high-level problem in computer vision is to differentiate motions of bodies in a video sequence. It is an essential building block for purposes such as traffic monitoring, moving object segmentation and motion analysis. In general, motion can be considered to be a transformation of an object in space and time. This transformation could be represented using epipolar geometry as a projective transformation between two views \cite{epipolargeometry} or any other parametric representation. In recent years, many algorithms have been developed to segment motion, they can be divided into two main groups - Multiview-based and Two-view-based.

The Input for the Multiview-based approaches are Point trajectories \cite{14, 2, 35, GPCA, iclm_art, LatLRR, multicut, PAC, SSC}. Point trajectories are subsets consisting of selected feature points that are tracked along a sequence. Point trajectories, can be considered as a representation of motion of objects moving in the scene. In order to obtain a segmentation result, Multiview-based methods primarily use local information around each trajectory in order to create affinities between trajectories, which can subsequently be segmented using spectral clustering techniques \cite{35}. Multiview-based methods generally work well and are robust to noise because they handle the entire sequence at once. These methods usually compute pairwise affinities between all detected features over all frames. Consequently, complicated scenarios are accompanied by a high computational cost. 

The alternate fundamental approach for motion segmentation are Two-view-based methods. These methods \cite{10, 13, 20, 7, randomVoting, 16} ], usually focus on the epipolar constraint. Although these approaches are efficient, they only process a small part of the information available \cite{10, 13, 20, 7, randomVoting, 16}. For example, an object that had moved slowly or did not move significantly between two frames, would not be differ easily from the background. A further constraint on these methods is the use of random initialization. In practice, this means that convergence is not guaranteed every time and usually its labeling result is not-consistent. 

In this paper, we propose an initialization stage for the state-of-the-art algorithm presented at CVPR called Random Voting (RV) \cite{randomVoting}. We focus on extending its performance to deal with the following issues:
\begin{description}
  \item[$\bullet$ ] Expedite convergence - as RV is an iterative algorithm, performance depends on the number of iterations for convergence 2
  \item[$\bullet$ ] Minimize accuracy distribution rates - even though accuracy can be high, it can vary as random initialization by RV \cite{randomVoting} yields non-consistent accuracy rates 
 \item[$\bullet$ ] Improve accuracy with noise-free data and increase robustness to noise 
\end{description}
 
By employing an algorithm based on RV \cite{randomVoting} and focusing on the above issues, we have developed the Locally Affine Atom Voting (LAAV) method. The LAAV is a fast and accurate motion segmentation algorithm that incorporates two stages. The first stage is a pre-processing stage based on a novel representation of objects called Piecewise-Smooth. The pre-processing stage provides a hard-assignment for motion features, followed by a pair-wise fine-tuning using RV in the segmentation stage.
As a result of the initialization stage, the RV algorithm becomes more robust to complex scenarios, is more accurate and less variant.  Our algorithm's robustness to noise is comparable to other state-of-the-art algorithms.

The paper is organized as follows: Section \ref{sec:Related-Work} discusses relevant past work. In section \ref{sec:Motion-Segmentation-Using} we present LAAV method and discuss its flow and components. Section \ref{sec:Results} describes experiments done: comparison of performance and an extensive evaluation of LAAV in conjunction with leading state-of-the-art RV. We conclude our findings in section \ref{sec:Conclusions}.

\section{Related Work} \label{sec:Related-Work}
As described above, there are two major groups of approaches to motion segmentation: Multiview and Two-view. 

The Two-view approach is derived merely from the relative camera poses from multiple views, called relative-pose constraints, without any additional assumptions of the scene. The epipolar constraint is such a constraint between two views \cite{Luong1996}. 

Random Voting (RV) \cite{randomVoting}, which is considered as the leading geometric method for motion segmentation partly because of its robustness to noise, has shown particularly successful results with a low computational cost. The algorithm, based on epipolar geometry, is an iterative process of randomized feature selection between two frames, estimating a fundamental matrix from the selected features and vote scores for the rest of the remaining features to be associated with a certain motion model. Since the method uses random initialization, it never loses any information even when the selected features do not represent a motion model. However, this approach only works well when the independent moving object is big enough, such that it consists of enough features to properly estimate the object's motion. In addition, objects in the scene need to be in a certain size so that the background/object features ratio is not too high in order for the object's features to be selected in the randomized features selection. Finally, its accuracy rate results can vary due to the random initialization.  

 The Multiview approach utilizes the trajectory of the feature points. PAC \cite{PAC} and SSC \cite{SSC} methods have quite accurate results in multiple motion cases in a sequence and are also robust to noise. However, those algorithms are extremely slow. Latent low-rank representation-based method (LatLRR) \cite{LatLRR} is faster and more accurate, but this method becomes degraded in extremely noisy environments. The ICLM-based approach \cite{iclm_art} is very fast, but has lower accuracy than other state-of-the-art approaches. In addition, while Multiview approaches are more accurate than Two-view approaches, they do not have good performance when there are only a few frames.
\begin{figure}
\setlength{\belowcaptionskip}{-8pt}
\begin{center}
\begin{tabular}{ccc}

\bmvaHangBox{{\parbox{2.7cm}{~\\[0mm]
 \hspace{2.24mm}\includegraphics[width=2.8cm]{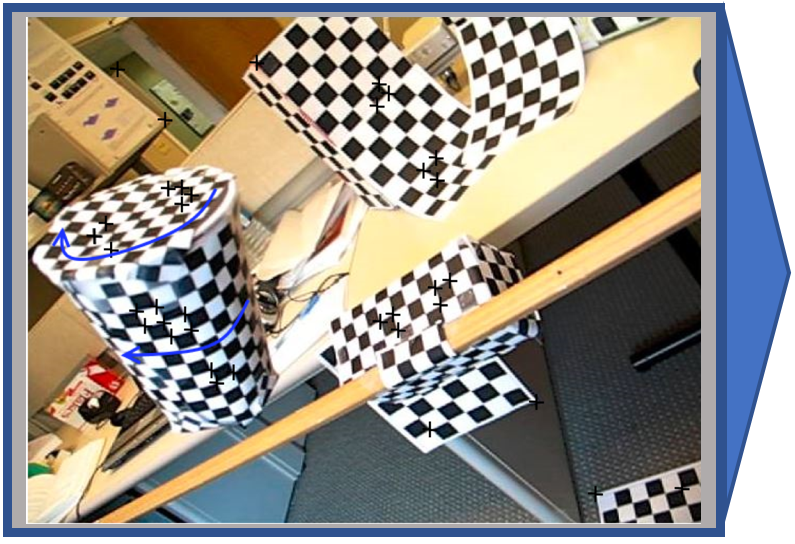}}}}&
\bmvaHangBox{{\includegraphics[width=2.8cm]{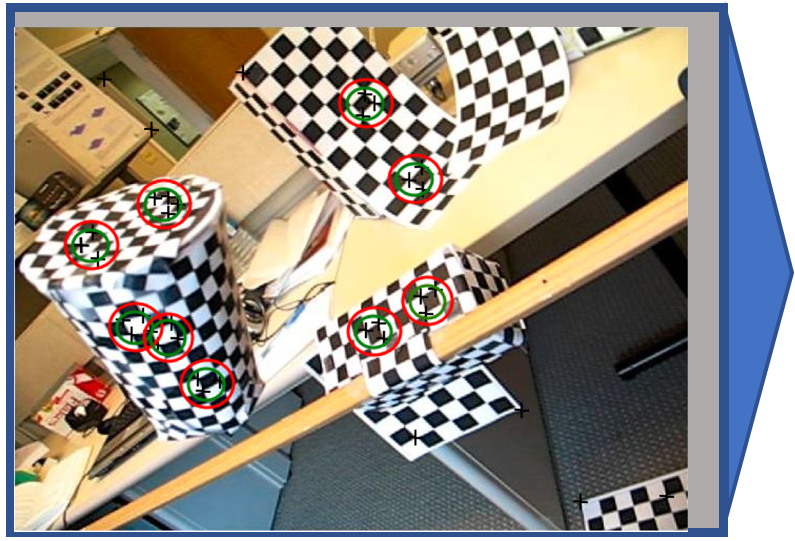}}}&
\bmvaHangBox{{\includegraphics[width=2.8cm]{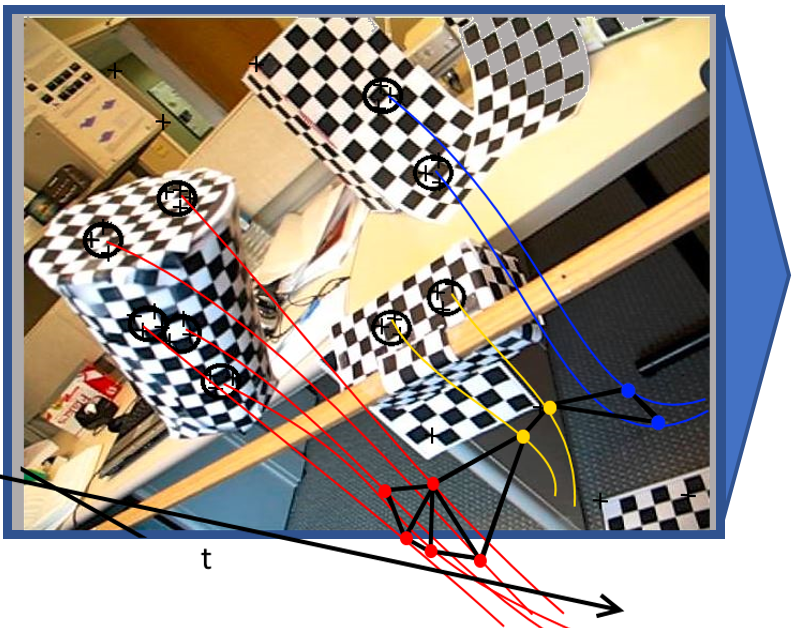}}}\\
(a)&(b)&(c) \\
\bmvaHangBox{{\includegraphics[width=2.8cm]{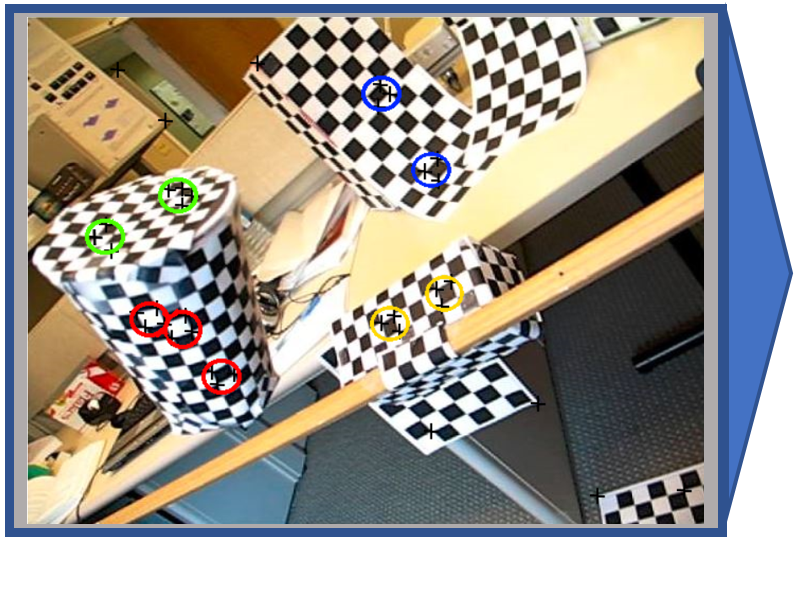}}}&
\bmvaHangBox{{\includegraphics[width=2.8cm]{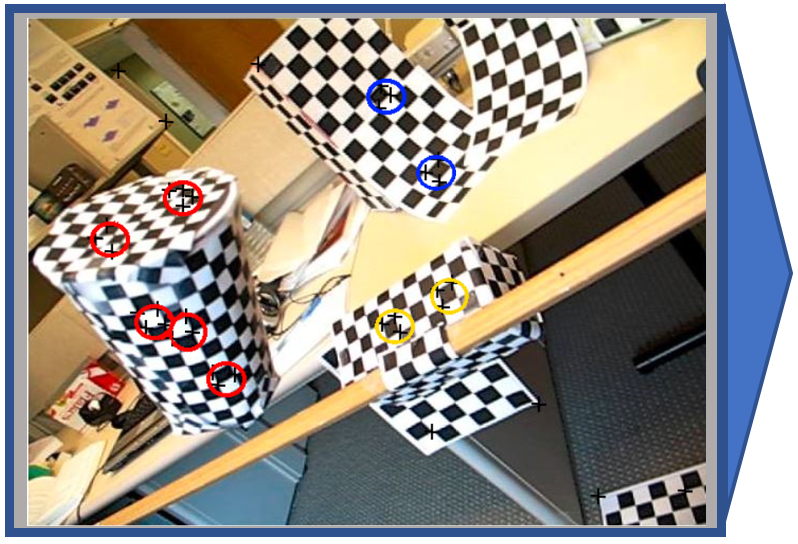}}}&
\bmvaHangBox{{\includegraphics[width=2.8cm]{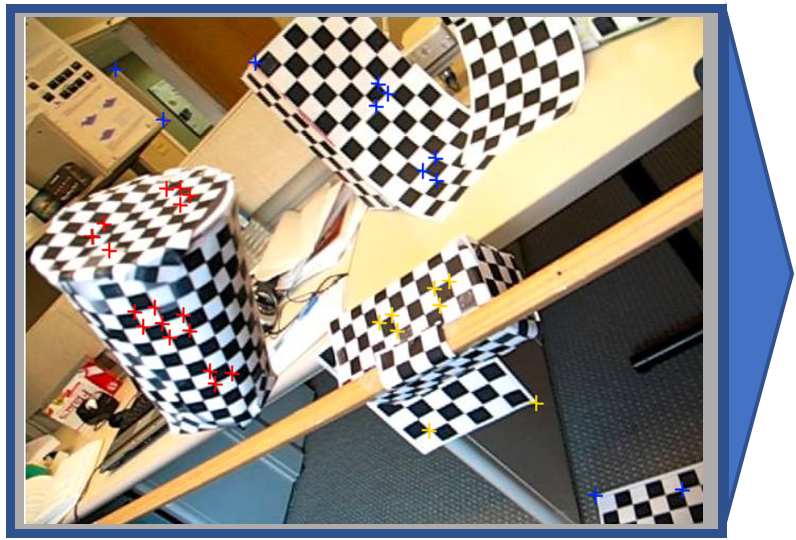}}}\\
(d)&(e)&(f) 
\end{tabular}
\end{center}
\caption{Overall flow of the proposed method. (a) The Video obtained from \cite{hopkins155} with its feature points detected and tracked along the sequence. (b) Visualization of atoms structure obtained using object's piece-wise smooth representation, all feature confined by the red circle would get joined to an atom. (c) shows segmentation results based on atoms trajectories as a minimum cost multicut problem.  (d)  shows LAAV output result with over segmented independent objects. (e) Fine-To-Coarse output stage; groups of the same moving object are joined for an object motion segmentation. (f) Final segmentation after a moderate Random Voting(RV) to segment features that had not been joined to an atom. }
\label{fig:Conecpt-of-the}
\end{figure}

\section{Motion Segmentation Using LAAV} \label{sec:Motion-Segmentation-Using}
In this section, we discuss the structure of the proposed algorithm comprised from blocks presented in Figure \ref{fig:Conecpt-of-the}. 

Using the proposed method, all tracked 2D feature points along a video sequence are segmented into either background or independent moving objects. We denote a 2D feature point in a single frame $l$ as $y_k^l$ where $k=1,.,.,K$, with $K$ as the total number of features and $l=1,…,L$, with $L$ as the total number of frames. $y_k^l$  is represented in homogeneous coordinates. 
 
{\bf Objects Piecewise-Smooth Representation} -  We present a novel representation of an object's motion which relies on the fact that object regions are often piecewise-smooth, i.e. can be represented as a linked overlapped patches (referred to as atoms) with gradually differing affine transformations. By explicitly incorporating the unique object's piecewise-smooth quality, atoms are robustly estimated from the matched feature points. Unlike standard approaches for motion segmentation, the presented algorithm uses atoms instead of pairwise affinities between all features;  thus, significantly simplifies complex scenarios and reduces the computational cost without the loss of accuracy. Using this representation, we can estimate a fine \cite{fine_grained} segmentation of the motion models in the scene. i.e. non-rigid bodies could hold multiple motion models.

The LAAV input is a set of all feature points that are tracked along the sequence and their trajectories.

In order to simplify the complexity of the scenario, neighbored feature trajectories sharing similar affine transformations are joined to atoms denoted as $x_i$ where $i=1, . . . ,N$ and $N$ is the total number of atoms. The set $X={ x_1   ,x_x  ,...,x_N  }$ consists of all atoms in a sequence. Figure \ref{fig:Conecpt-of-the} (b) visualizes the constructed atoms in their designed structure, all features inside the red circle belong to an atom and could be joined to other atoms while features inside the green circle are only belong to a single atom.

{\bf Fine 2D motion estimation } -  Non-rigid bodies can consist of multiple independent parts moving in different directions. Therefore, the number of motion models in every sequence could be considered as bigger than the number of the independent moving objects.  In this stage, each atom along a video sequence is classified into either background or independent motion, resulting in a fine-grained \cite{fine_grained} description of all the independent moving objects in the sequence. For that purpose we formulate the segmentation of a video sequence based on atom trajectories as a minimum cost multicut problem \cite{multicut}, as shown in Figure \ref{fig:Conecpt-of-the} (c). Next, the set of atoms $X={ x_1   ,x_x  ,...,x_N  }$  are segmented to an over-estimate of motion models when given the true number of independent moving objects in the sequence, denoted as C. We found empirically that the number $2C$ of motion models is suitable for over-estimation of all motion models under the Hopkins155 dataset \cite{hopkins155} which was used to evaluate our algorithm. 

The LAAV result is the set $X$ and is a fine-grained description of all independent moving objects in the sequence. Each atom $x_i$  in the set $X$ is being labeled as $S_c$ where $c = 1,...,2C$. i.e. $S_c$ express motion models labels or the set of feature trajectories belonging to the $c^{th}$ motion model. Figure \ref{fig:Conecpt-of-the} (d) shows the segmentation results of LAAV.

{\bf Fine-To-Coarse 2D motion estimation }  - In order to reconstruct independent motions models to independent moving objects the next stage is applied. Here, all independent motions are joined in order to describe motion in a coarse-grained \cite{fine_grained} mode. i.e. to assemble independent moving objects from all motion models. The $2C$ motion models from previous stage are joined down to $C$ motion models. Figure \ref{fig:Conecpt-of-the} (e) shows the segmentation results of the Fine-To-Coarse procedure resulting with the set $X$ being labeled as $S_c$. 

{\bf Fine-tuning Randomized Voting} - Finally, this stage utilizes a moderate version of RV \cite{randomVoting} in order to label features that had not yet been joined to atoms marked in black in Figure \ref{fig:Conecpt-of-the}(a)-(e), and were not segmented to any independent moving object. 
Figure \ref{fig:Conecpt-of-the} (f) shows the final motion segmentation results.

\subsection{Atom Construction}
Perspective projection is often used in order to model motion of rigid objects between pairs of frames. However, alternative geometric relationships that facilitate parameter computation have also been proven useful for this purpose \cite{doi:10.1137/080732730}. For instance, in small patches within images the perspective transformation can effectively be approximated by an affine transformation \cite{doi:10.1137/080732730}. Furthermore, examining small patches of non-rigid bodies can be worthwhile such that any non-rigid body can be represented as a combination of multiple small rigid bodies.

Embracing this representation, we can define the term piecewise-smooth object. Piecewise-smooth objects are non-rigid bodies where the temporal transformation between two frames are often slow in small regions. On piecewise-smooth objects, the affine transformation for neighboring patches is similar. Hence, a group of small patches where all features correspond to the same affine transformation can be grouped and denote as atoms. 

We define atoms as feature sets that comply with the following conditions: 

{\bf Condition 1}  {\em for every atom $x_i \in X$ the minimum number of features should be no less than 3}

Affine transformation has 6 degrees of freedom; translation + rotation + scale + aspect ratio + shear, so in order to estimate an affine transformation, atoms should consist of a minimal 3 features. The atoms construction is as follows: atoms are constructed sequentially. Iteratively, two frames $l,r$ are selected randomly, with enough frame separation so that a minimum movement can be detected. Then, a feature is randomly selected and a minimum of 3 neighbors are located. 

{\bf Condition 2}  {\em each atom $x_i$ should be comprised out of a set of features $Y$, bounded by the external radius $R_2$} 

The atom's bounding area is imperative so that the affine transformation approximation is maintained. In order to reduce computational cost, we denote an additional bounding radius $R_1$. All features that had been joined to an atom, with a Euclidean distance from the origin less than $R_1$ are excluded from being joined to other atoms in the followed iterations. Next, the set of features $Y$ is tested for consensus using RANSAC \cite{RANSAC} under affine transformation. If there is a consensus, an atom is created. The process continues for the rest of the features that had not yet been joined to an atom. There is a trade-off for selecting the radius $R_2$: A radius that is too big means more atoms will be comprised, but the affine transformation premise will no longer be valid; choosing a $R_2$ radius that is too small will comply under the affine transformation premise, but fewer atoms will be comprised.

{\bf Condition 3} {\em $x_i \cap x_j  \neq \emptyset$, Where $X$ is a joint set where atoms overlap }

Condition 3 is imperative so that independent motion objects can be represented as piecewise-smooth objects. When applying overlap between atoms we can represent the transformation of a piecewise-smooth object as a minimum cut problem \cite{multicut}. The minimum cut problem is the problem of decomposing a graph $G=( V,E )$ into an optimal number of segments such that the overall cost in terms of edge weights $w_e$ is minimized. 

This node labeling problem can equivalently be formulated as a binary edge labeling problem 
 
\begin{align}
\underset{x\in\left\{ 0,1\right\} ^{E}}{min}\sum_{e\in E}w_{e}x_{e}\\
subject\,to\,x\in MinimumCut\nonumber 
\end{align}
 
We build the graph $G$ such that every atom is represented by a vertex $v \in V$. If every vertex is connected by an edge $e \in E$ to its nearest neighbors, all solutions to the minimum cut problem yield a segmentation into connected components. The weights of the edges $e \in E$ define how similar two atoms are. We determine atom affinities using the forward-backward error evaluated using the affine transformation of an atom between two frames on a different atom. 

The minimum cost multi-cut computed on $G$ yields our desired segmentation into the optimal number of motion models. Since objects consist of multiple independent moving parts, the optimal number of segments is selected to be larger than the true number. The process yields a fine-grained segmentation later to be coarsened.

The fine-grained segmentation result by LAAV show the advantage of the presented approach. LAAV is not restricted to just a rigid body segmentation algorithm but also can describe more complicated motions. For example in Figure \ref{fig:Conecpt-of-the} (c), the motion of the cylinder's upper face and middle face are segmented separately. We later coarsen the segmentation results for the purpose of comparison with other state-of-the-art algorithms and to enhance run-time.

\subsection{Fine-To-Coarse}
To merge the atoms, we suggest a Fine-To-Coarse inference strategy that utilizes motion segmentation with epipolar geometry - the intrinsic projective geometry between two views. The fundamental matrix F represents the epipolar geometry, which satisfies the following condition:
\begin{equation}
y_{2}^{T}Fy_{1}=0 \label{eq:epipolarGeomtry}
\end{equation}

Where the fundamental matrix $F$ is a 3x3 matrix of rank 2, $y_1$ and $y_2$ are features in homogeneous coordinates;  $y_{1}=[y_{1_{x}},y_{1_{y}},1]^{T}$ ; and $y_{2}$ = $[y_{2_{x}}',y_{2_{y}}',1]^{T}$. under Equation \ref{eq:epipolarGeomtry} we can infer the following properties: 

{\bf Property 1} all atoms on the same moving rigid object have the same fundamental matrix $F$. 

{\bf Property 2} the corresponding feature lies on a moving object in the other frame and must lie on the epipolar line. Where $F_{y_1}$ represents an epipolar line in the other frame and $y_2$ lies on the line. 

In Fine-To-Coarse, we adopt a scoring scheme of all motion models estimated in the previous step. Our approach randomly selects 2 atoms from one motion model $X_{S_{j}}=\left\{ x_{j1},x_{j2}\in S_{j}\right\} $ and 2 atoms from another motion model $X_{S_{k}}=\left\{ x_{k1},x_{k2}\in S_{k}\right\} $  and $j\ne k$. Under Property 1, if both motion models are on the same moving body their fundamental matrix $F$ should be the same. We estimate the fundamental matrix $F$ from the 4 atoms. Once the fundamental matrix $F$ is estimated, we use the Sampson distance to measure the distance between atoms and the epipolar line. Under Property 2, a small distance between a corresponding feature to the epipolar line means that the fundamental matrix $F$ represents the motion model accordingly. In order to evaluate the Sampson distance on atoms we measure the distance of all features comprising the atoms and the epipolar line. We use it as follows:

\begin{equation}
SD(X_{S_{j}},X_{S_{k}},F_{S_{j},S_{k}})=\frac{X_{S_{j}}^{T}F_{S_{j},S_{k}}X_{S_{k}}}{(F_{S_{j},S_{k}}FX_{S_{j}})^{2}+(F_{S_{j},S_{k}}X_{S_{k}})^{2}}
\end{equation}

We measure the distance $d_{j,k}=SD(X_{S_{j}}^{(l)},X_{S_{k}}^{(r)},F_{S_{j}S_{k}}^{(l,r)})$ between the epipolar line from the matrix $F_{S_{j}S_{k}}^{(l,r)}$, and two corresponding atoms between frames ( $l,r$ ) from the motion models $s_k$ and $s_j$ . The distance between the two motion models can be used to vote for two motion models belonging to the same motion. Therefore, every atom's $x_{i}\in\{X_{S_{j}},X_{S_{k}}\}$ distance from the epipolar line is voted by $e^{-\lambda d_{i}}$  where the parameter $\lambda$ controls the voting strength. If the value of $\lambda$ is small, then it gives a large voting value to the atoms and vice-versa. As a result of this process, $d_{j,k}$ is now a vector containing the distances from all features points in frame $r$ to the epipolar line. 

\begin{equation}
d_{j,k}^{epipolar}=max(avg(d_{j,k}),avg(d_{k,j}))\,
\end{equation}

$d_{j,k}^{epipolar}$ epipolar is the total distance between the atoms chosen randomly. In other words, if the distance value is large, the affinity between the two selected motion models is small. In contrast to \cite{randomVoting} we do not change the motion model labeling during accumulation, only at the end of the process. Since we do not deal with pairwise labeling, we do not need to worry about misclassification. In addition, this process is not iterative, only a small number of 7 frames are selected. The frames are selected randomly with enough distance between them so that different motions will be well separated. 

We also compute motion models' affinities determined by the spatial distance between trajectories. Similarly, to \cite{multicut}, we define the motion difference of two motion models at time $t$ as $d_{t}^{motion}(A,B)=\frac{\Vert\partial A-\partial B\Vert}{\sigma_{t}}$, where $\partial_{t}A$ and $\partial_{t}B$ are the partial derivatives of the averaged features trajectories of atoms $A$ and $B$. $A$, $B$ belong to the motion model $j,k$ respectively. $\sigma_{t}$ is the variation of the optical flow \cite{5206697}. We build this problem as a superposition of both the spatial distance and the epipolar distance.

\begin{equation}
Z_{j,k}=Z_{j,k}+exp(-\lambda\cdot d_{j,k}^{epipolar}\cdot d_{t}^{motion}(S_{j},S_{k}))^{-1}
\end{equation}

where $\lambda$ is a forgetting-factor controlling the weight of every sample. In order to group the motion model, we utilize a spectral clustering method \cite{NIPS2001_2092} on $Z$, and $Z$ is the $(2C\,x\,2C)$  affinity matrix. 
\subsection{Fine Tuning}
The Fine-Tuning stage is designated to segment sparse features that had not been segmented to an atom and consequently was not segmented to a motion model. To this end we use the randomized voting (RV) method \cite{randomVoting} which performs the best up-to-date motion segmentation and is robust to noise. RV also requires prior knowledge of the motion models $C$. In contrast to RV, we do not use random initialization for the groups, rather use the features that had been labeled in LAAV and Fine-To-Coarse stages for initialization. At this stage there is a high confidence in the features labeling, therefore an additional weight is added to the voting histogram so that RV will affect the unlabeled ones more than the labeled ones. The algorithm is similar to RV; for all groups the fundamental matrices are estimated using the initial labels. Next, we update the labels using the fundamental matrices estimated from the current features segmented to the group, and the process is repeated. For each iteration the algorithm selects $m$-features and estimates the fundamental matrix $F$ using the selected features. Then, the Sampson distance is calculated using the matrix $F$ for all the features. A score is added to the histogram for a feature as a function of the Sampson distance from the epipolar line. Finally, the labels of the features are updated, where labels are assigned according to the maximum score.
However, the randomized voting could drift to an incorrect result due to bad fitting of the forgetting-factor $\alpha$, or a small voting strength parameter $\lambda$. In this case, we use a large voting strength and forgetting-factor. Due to using these parameters we do need to worry about termination conditions, because the prior labeling was complete with the exception of minor fine-tuning.
\section{Results} \label{sec:Results}
We evaluated our algorithm using Hopkins155 dataset \cite{hopkins155}. The Hopkins155 dataset has been the standard evaluation metric for the problem of motion segmentation. It contains 155 sets of features point trajectories of two and three motion models from 50 videos, along with the corresponding ground truth segmentation for comparison. We compared our proposed algorithm's performance with the state-of-the-art methods with emphasis on \cite{randomVoting} as it showed best results so far. 
For evaluating the quality of the methods, we have two crucial criteria, the accuracy with noise-free data and with added artificial noise. The following methods were compared: generalized principal component analysis (GPCA) \cite{GPCA}; SCC presented in \cite{scc}; The spectral clustering (SC) method \cite{SC}; Ranking of locally sampled subspaces (RLS) \cite{RLS}; Linear combination of views based algorithm(LCV)\cite{LCV} ; SSC \cite{SSC}; PAC \cite{PAC}; Random Voting(RV)\cite{randomVoting};LatLRR \cite{LatLRR}, and ICLM \cite{iclm_art}.

\begin{table}
\bmvaHangBox{ 
\includegraphics[width=1\textwidth]{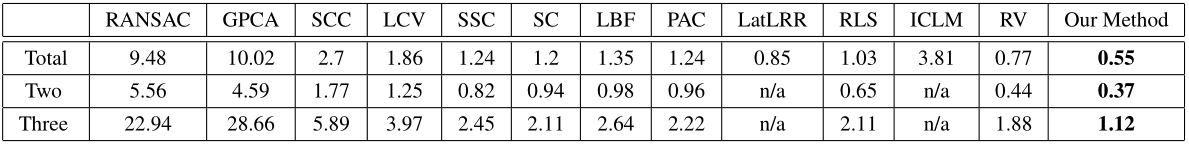}} 
\caption{Total Errors. The first row represents average error rates for all (155 sequences). The second and third rows are two (120 sequences) and three (35 sequences) motions, respectively. n/a means the value is not presented in the corresponding paper.}
\label{tab:results_comparison}
\end{table}
\begin{figure}
\setlength{\belowcaptionskip}{-10pt}
\begin{minipage}[b]{.32\textwidth} \label{fig:accuracy}
\centering
 
\bmvaHangBox{ 
\includegraphics[width=1.13\textwidth]{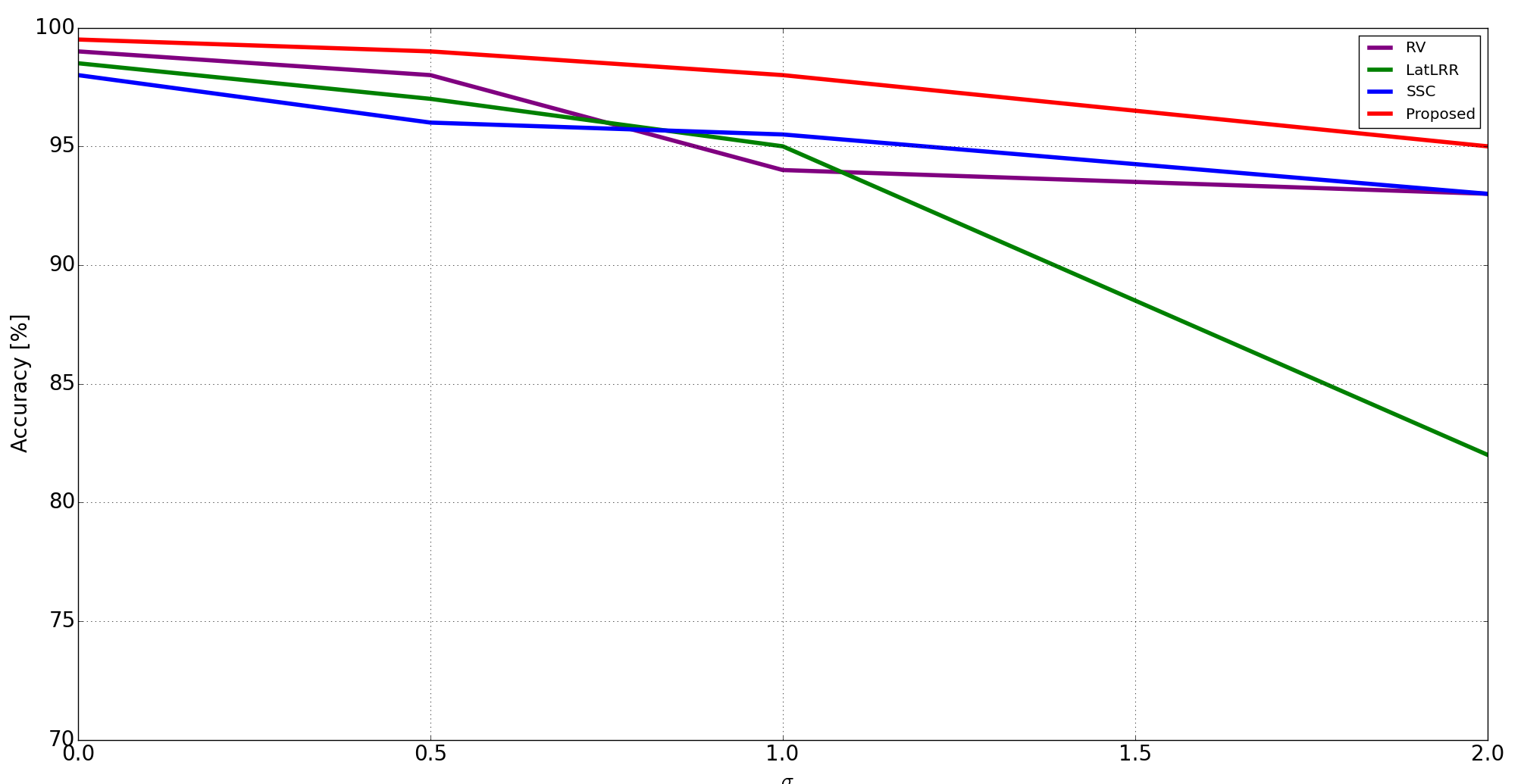}}
\caption{Accuracy with Artificial Noise using Hopkins155 with different levels of Gaussian noise}

\end{minipage}
\hfill
\begin{minipage}[b]{.31\textwidth} \label{fig:Number-Of-Iteration-1}
\centering
\bmvaHangBox{{\parbox{4.0cm}{~\\[-1mm]
\includegraphics[width=.9\textwidth]{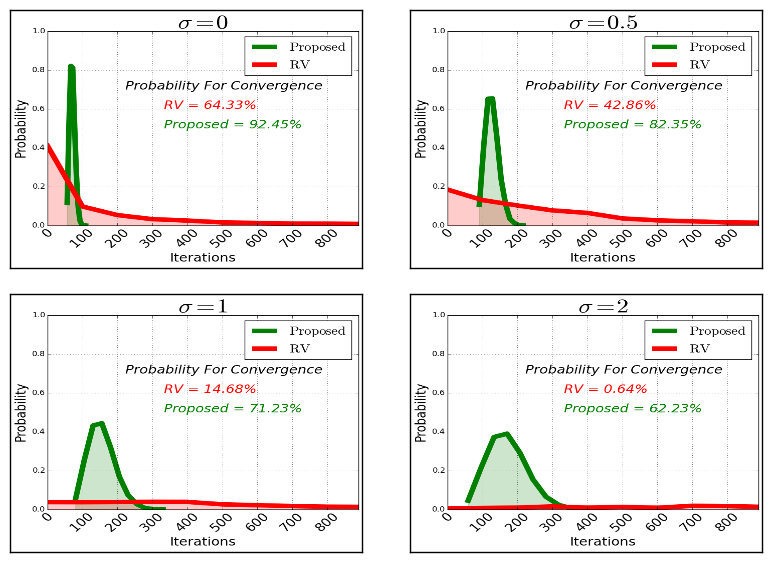}}}} 

\caption{Number Of Iteration for Accurate motion segmentation for $\sigma_{n}\in\left\{ 0,0.5,1,2\right\}$ . }
\end{minipage}
\begin{minipage}[b]{.31\textwidth} \label{fig:Accuracy-Distribution}
\centering
\includegraphics[width=.9\textwidth]{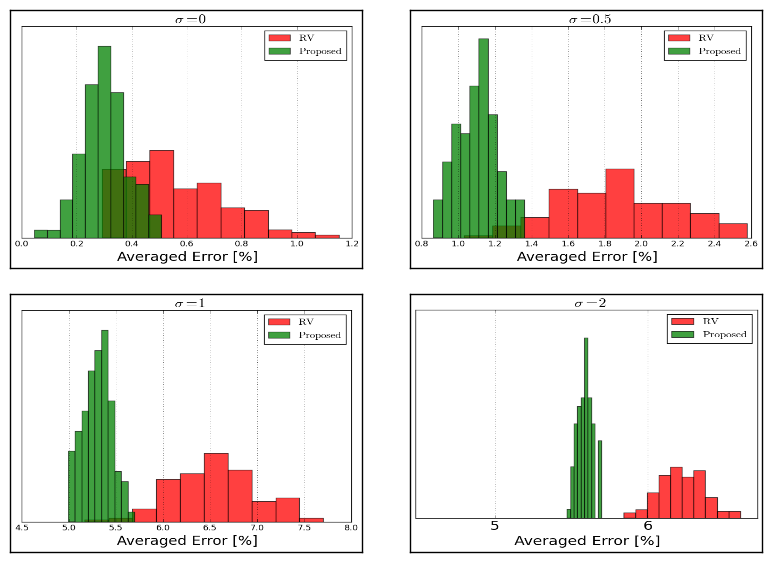}
\caption{Accuracy Distribution for both methods for $\sigma_{n}\in\left\{ 0,0.5,1,2\right\} $}
\end{minipage}
\end{figure}
\subsection{Performance Comparison}
In Table \ref{tab:results_comparison} we present the average misclassification error over two-motions, three-motions, and both. In all three cases, the performance of our algorithm showed the lowest error rate when compared to the other algorithms. Our algorithm achieved total misclassification average error of 0.55\%. In two-motions and three-motions separately we achieved 0.37\% and 1.12\% respectively. 
Another experimental setup is designated to test the effects of noise of different magnitudes on the classification accuracy. Following \cite{randomVoting}, we used three Gaussians noise with a zero mean and different diagonal covariance matrix  $\sum_{n}=\sigma_{n}^{2}I\:,\sigma_{n}\in\left\{ 0.5,1,2\right\} $ ,where, $I$ is a $2X2$ identity matrix. In the test we computed over 150 repetitions. Results are shown in Figure \ref{fig:accuracy}, and indicate that our algorithm showed a comparable performance to the other state-of-the-art methods. Like other algorithms, large noise variances comprise classification accuracy, and so, in the same manner the contribution of LAAV is limited. Large noise variances result in only a small portion of atoms constructed, leaving a large portion of features non-initialized. In this scenario the algorithmic approach is to randomly initialize the remained features as in \cite{randomVoting}.

It can be seen that, from all the state-of-the-art methods, the iterative algorithm RV outperforms all. The result of the randomized voting is reliable when the number of parameters for estimation is small and the amount of feature points is large. Thus, convergence can be considered as definite. However, if in other cases the process does not converge the results are less reliable, for example, when the number of features is too large. We show that our proposed method consistently outperforms RV on Hopkins155 dataset in Table \ref{tab:results_comparison} and in Figure \ref{fig:accuracy}.
\subsection{Proposed vs. Random Voting}
Two criteria are relevant when comparing iterative methods, the accuracy performance and variance, and the process convergence, it's distribution as a function of iteration number and it's probability. Using the Hopkins155 \cite{hopkins155} dataset we compared the distribution of the iteration number when the process converged between our proposed method in the Fine-Tuning Step and RV. Figure \ref{fig:Number-Of-Iteration-1} shows the distribution of the iteration number for convergence and convergence percentage under different levels of artificial noise. In order to generate statistical significance, we repeated the test 150 times both for noise-free data and for added artificial noise. Following \cite{randomVoting} we used three Gaussians noise with a zero mean and different diagonal covariance matrix $∑_n$. In the RV algorithm, the iteration termination criteria is set by the number of trials. In RV all trials are independent and under the assumption that for enough trials the algorithm is bounded to converge. RV process is divided into trials and iterations. Every trial consists of multiple iterations. A trial is terminated when the iteration termination criteria is reached, then another trial starts. The calculation of RV iteration number was performed as follows: for every trial, following \cite{randomVoting} the maximum iterations termination criteria by default is 150. For every trial that does not converge the iteration number increases by 150. Following \cite{randomVoting} The maximum trial termination criteria by default is 10. We excluded sequences that did not converge and estimated the ratio between the converged sequence with the ones that did not converge. It can be seen from Figure \ref{fig:Number-Of-Iteration-1} that our method converges with significantly less iterations. Also, in all noise levels the number of non-converged sequences overall in our approach is far smaller.

In order to compare the accuracy variation between RV and the proposed method we repeated the test 150 times for each of the non-contaminated or contaminated data with three Gaussians noise with a zero mean and different diagonal covariance matrix $∑_n$.  Results are shown in Figure \ref{fig:Accuracy-Distribution}. We show the accuracy percentage error with 4 different values of added noise. The Gaussians shown indicate that our method comprises accuracy for large levels of noise, while still showing better accuracy and small variation than RV.

\section{Conclusions} \label{sec:Conclusions}
We introduced a computationally efficient motion segmentation algorithm for trajectory data. Efficiency comes from the use of a simple but powerful representation of motion as atom trajectories, built from small patches of objects. We showed that an effective initialization for RV based on affine transformation of atoms is more accurate and robust to noise. As a result, our algorithm achieved excellent performances with total average error rates of 0.55\%. In conclusion, our algorithm achieved, within a reasonable time, the highest performance of all other state-of the art algorithms, and also achieved comparable accuracy under noisy environmental conditions.


\begin{thebibliography}{27}
\providecommand{\natexlab}[1]{#1}
\providecommand{\url}[1]{\texttt{#1}}
\expandafter\ifx\csname urlstyle\endcsname\relax
  \providecommand{\doi}[1]{doi: #1}\else
  \providecommand{\doi}{doi: \begingroup \urlstyle{rm}\Url}\fi

\bibitem[Brox et~al.(2009)Brox, Bregler, and Malik]{5206697}
T.~Brox, C.~Bregler, and J.~Malik.
\newblock Large displacement optical flow.
\newblock In \emph{2009 IEEE Conference on Computer Vision and Pattern
  Recognition}, pages 41--48, June 2009.
\newblock \doi{10.1109/CVPR.2009.5206697}.

\bibitem[Chen and Lerman(2009)]{scc}
Guangliang Chen and Gilad Lerman.
\newblock Motion segmentation by {SCC} on the hopkins 155 database.
\newblock \emph{CoRR}, abs/0909.1608, 2009.
\newblock URL \url{http://arxiv.org/abs/0909.1608}.

\bibitem[Costeira and Kanade(1998)]{2}
Jo\~{a}o~Paulo Costeira and Takeo Kanade.
\newblock A multibody factorization method for independently moving objects.
\newblock \emph{Int. J. Comput. Vision}, 29\penalty0 (3):\penalty0 159--179,
  September 1998.
\newblock ISSN 0920-5691.
\newblock \doi{10.1023/A:1008000628999}.
\newblock URL \url{http://dx.doi.org/10.1023/A:1008000628999}.

\bibitem[Dimitriou and Delopoulos(2012)]{RLS}
N.~Dimitriou and A.~Delopoulos.
\newblock Improved motion segmentation using locally sampled subspaces.
\newblock In \emph{2012 19th IEEE International Conference on Image
  Processing}, pages 309--312, Sept 2012.
\newblock \doi{10.1109/ICIP.2012.6466857}.

\bibitem[Elhamifar and Vidal(2012)]{SSC}
Ehsan Elhamifar and R.~Vidal.
\newblock Sparse subspace clustering: Algorithm, theory, and applications.
\newblock \emph{CoRR}, abs/1203.1005, 2012.
\newblock URL \url{http://arxiv.org/abs/1203.1005}.

\bibitem[Fischler and Bolles(1981)]{RANSAC}
Martin~A. Fischler and Robert~C. Bolles.
\newblock Random sample consensus: A paradigm for model fitting with
  applications to image analysis and automated cartography.
\newblock \emph{Commun. ACM}, 24\penalty0 (6):\penalty0 381--395, June 1981.
\newblock ISSN 0001-0782.
\newblock \doi{10.1145/358669.358692}.
\newblock URL \url{http://doi.acm.org/10.1145/358669.358692}.

\bibitem[Flores-Mangas and Jepson(2013)]{iclm_art}
Fernando Flores-Mangas and Allan~D. Jepson.
\newblock Fast rigid motion segmentation via incrementally-complex local
  models.
\newblock In \emph{CVPR}, pages 2259--2266. IEEE Computer Society, 2013.

\bibitem[Jian and Chen(2010)]{7}
Yong-Dian Jian and Chu-Song Chen.
\newblock Two-view motion segmentation with model selection and outlier removal
  by ransac-enhanced dirichlet process mixture models.
\newblock \emph{International Journal of Computer Vision}, 88\penalty0
  (3):\penalty0 489--501, 2010.
\newblock ISSN 1573-1405.
\newblock \doi{10.1007/s11263-010-0317-y}.
\newblock URL \url{http://dx.doi.org/10.1007/s11263-010-0317-y}.

\bibitem[Jung et~al.(2014)Jung, Ju, and Kim]{randomVoting}
Heechul Jung, Jeongwoo Ju, and Junmo Kim.
\newblock Rigid motion segmentation using randomized voting.
\newblock In \emph{Computer Vision and Pattern Recognition (CVPR), 2014 IEEE
  Conference on}, June 2014.
\newblock \doi{10.1109/CVPR.2014.158}.

\bibitem[Keuper et~al.(2015)Keuper, Andres, and Brox]{multicut}
M.~Keuper, B.~Andres, and T.~Brox.
\newblock Motion trajectory segmentation via minimum cost multicuts.
\newblock In \emph{IEEE International Conference on Computer Vision (ICCV)},
  2015.
\newblock URL
  \url{http://lmb.informatik.uni-freiburg.de//Publications/2015/KB15b}.

\bibitem[Krause et~al.(2013)Krause, Stark, Deng, and Fei-Fei]{fine_grained}
J.~Krause, M.~Stark, J.~Deng, and L.~Fei-Fei.
\newblock 3d object representations for fine-grained categorization.
\newblock In \emph{2013 IEEE International Conference on Computer Vision
  Workshops}, pages 554--561, Dec 2013.
\newblock \doi{10.1109/ICCVW.2013.77}.

\bibitem[Lauer and Schnorr(2009)]{SC}
F.~Lauer and C.~Schnorr.
\newblock Spectral clustering of linear subspaces for motion segmentation.
\newblock In \emph{2009 IEEE 12th International Conference on Computer Vision},
  pages 678--685, Sept 2009.
\newblock \doi{10.1109/ICCV.2009.5459173}.

\bibitem[Li(2007)]{10}
H.~Li.
\newblock Two-view motion segmentation from linear programming relaxation.
\newblock In \emph{2007 IEEE Conference on Computer Vision and Pattern
  Recognition}, pages 1--8, June 2007.
\newblock \doi{10.1109/CVPR.2007.382975}.

\bibitem[Liu and Yan(2011)]{LatLRR}
Guangcan Liu and Shuicheng Yan.
\newblock Latent low-rank representation for subspace segmentation and feature
  extraction.
\newblock In \emph{2011 International Conference on Computer Vision}, pages
  1615--1622. IEEE, 2011.

\bibitem[Luong and Faugeras(1996)]{Luong1996}
Quan-Tuan Luong and Olivier~D. Faugeras.
\newblock The fundamental matrix: Theory, algorithms, and stability analysis.
\newblock \emph{International Journal of Computer Vision}, 17\penalty0
  (1):\penalty0 43--75, Jan 1996.
\newblock ISSN 1573-1405.
\newblock \doi{10.1007/BF00127818}.
\newblock URL \url{https://doi.org/10.1007/BF00127818}.

\bibitem[Morel and Yu(2009)]{doi:10.1137/080732730}
J.~Morel and G.~Yu.
\newblock Asift: A new framework for fully affine invariant image comparison.
\newblock \emph{SIAM Journal on Imaging Sciences}, 2\penalty0 (2):\penalty0
  438--469, 2009.
\newblock \doi{10.1137/080732730}.
\newblock URL \url{https://doi.org/10.1137/080732730}.

\bibitem[Ng et~al.(2002)Ng, Jordan, and Weiss]{NIPS2001_2092}
Andrew~Y. Ng, Michael~I. Jordan, and Yair Weiss.
\newblock On spectral clustering: Analysis and an algorithm.
\newblock In T.~G. Dietterich, S.~Becker, and Z.~Ghahramani, editors,
  \emph{Advances in Neural Information Processing Systems 14}, pages 849--856.
  MIT Press, 2002.
\newblock URL
  \url{http://papers.nips.cc/paper/2092-on-spectral-clustering-analysis-and-an-algorithm.pdf}.

\bibitem[Poling and Lerman(2013)]{13}
Bryan Poling and Gilad Lerman.
\newblock A new approach to two-view motion segmentation using global dimension
  minimization.
\newblock \emph{CoRR}, abs/1304.2999, 2013.
\newblock URL \url{http://arxiv.org/abs/1304.2999}.

\bibitem[Rao et~al.(2008)Rao, Tron, Vidal, and Ma]{14}
S.~R. Rao, R.~Tron, R.~Vidal, and Yi~Ma.
\newblock Motion segmentation via robust subspace separation in the presence of
  outlying, incomplete, or corrupted trajectories.
\newblock In \emph{2008 IEEE Conference on Computer Vision and Pattern
  Recognition}, pages 1--8, June 2008.
\newblock \doi{10.1109/CVPR.2008.4587437}.

\bibitem[Shi and Malik(2000)]{35}
Jianbo Shi and Jitendra Malik.
\newblock Normalized cuts and image segmentation.
\newblock \emph{IEEE Trans. Pattern Anal. Mach. Intell.}, 22\penalty0
  (8):\penalty0 888--905, August 2000.
\newblock ISSN 0162-8828.
\newblock \doi{10.1109/34.868688}.
\newblock URL \url{http://dx.doi.org/10.1109/34.868688}.

\bibitem[Torr(1998)]{16}
P.~H. Torr.
\newblock Geometric motion segmentation and model selection [and discussion].
\newblock \emph{Philosophical Transactions: Mathematical, Physical and
  Engineering Sciences}, 356\penalty0 (1740):\penalty0 1321--1340, 1998.
\newblock ISSN 1364503X.
\newblock URL \url{http://www.jstor.org/stable/54848}.

\bibitem[Tron and Vidal(2007)]{hopkins155}
R.~Tron and R.~Vidal.
\newblock A benchmark for the comparison of 3-d motion segmentation algorithms.
\newblock In \emph{2007 IEEE Conference on Computer Vision and Pattern
  Recognition}, pages 1--8, June 2007.
\newblock \doi{10.1109/CVPR.2007.382974}.

\bibitem[Vidal et~al.(2005)Vidal, Ma, and Sastry]{GPCA}
R.~Vidal, Yi~Ma, and S.~Sastry.
\newblock Generalized principal component analysis (gpca).
\newblock \emph{IEEE Transactions on Pattern Analysis and Machine
  Intelligence}, 27\penalty0 (12):\penalty0 1945--1959, Dec 2005.
\newblock ISSN 0162-8828.
\newblock \doi{10.1109/TPAMI.2005.244}.

\bibitem[Vidal et~al.(2002)Vidal, Soatto, Ma, and Sastry]{20}
Rene Vidal, Stefano Soatto, Yi~Ma, and Shankar Sastry.
\newblock Segmentation of dynamic scenes from the multibody fundamental matrix,
  2002.

\bibitem[Zappella et~al.(2010)Zappella, Provenzi, Llad{\'o}, and Salvi]{PAC}
Luca Zappella, Edoardo Provenzi, Xavier Llad{\'o}, and Joaquim Salvi.
\newblock Adaptive motion segmentation algorithm based on the principal angles
  configuration.
\newblock In \emph{Asian Conference on Computer Vision}, pages 15--26.
  Springer, 2010.

\bibitem[Zhang(1996)]{epipolargeometry}
Zhengyou Zhang.
\newblock Determining the epipolar geometry and its uncertainty: A review.
\newblock \emph{International Journal of Computer Vision}, 27:\penalty0
  161--195, 1996.

\bibitem[Zografos and Nordberg(2011)]{LCV}
Vasileios Zografos and Klas Nordberg.
\newblock Fast and accurate motion segmentation using linear combination of
  views.
\newblock In \emph{BMVC 2011 :}, pages 12.1--12.11, 2011.

\end{thebibliography}
\end{document}